\documentclass[11pt]{article}
\usepackage{booktabs} 
\usepackage{graphics}
\usepackage{epsfig,amsmath,amsthm,amsfonts,amssymb}
\usepackage{mathtools}
\usepackage{xcolor}
\usepackage{pgfplots}
\usepackage{algorithm}
\usepackage{algorithmic}
\usepackage{siunitx}
\usepackage[margin=.8in]{geometry}
\usepackage{graphicx}
\usepackage{subcaption}
\usepackage[shortlabels]{enumitem}
\usepackage{hyperref}

\usepackage[active]{srcltx}

\title{EmDT: Embedding Diffusion Transformer for Tabular Data Generation in Fraud Detection} 
\author{En-Ya Kuo\thanks{Arizona State University -- School of Mathematical and Statistical Sciences, Tempe, AZ, USA} \and Sebastien Motsch\footnotemark[1]}

\date{\today}

\begin{document}
\maketitle

\begin{center}
\small
This work has been submitted to the IEEE for possible publication. Copyright may be transferred without notice, after which this version may no longer be accessible.
\end{center}

\tableofcontents

\begin{abstract}
Imbalanced datasets pose a difficulty in fraud detection, as classifiers are often biased toward the majority class and perform poorly on rare fraudulent transactions. Synthetic data generation is therefore commonly used to mitigate this problem. In this work, we propose the Clustered \textbf{Em}bedding \textbf{D}iffusion-\textbf{T}ransformer (EmDT), a diffusion model designed to generate fraudulent samples. Our key innovation is to leverage UMAP clustering to identify distinct fraudulent patterns, and train a Transformer denoising network with sinusoidal positional embeddings to capture feature relationships throughout the diffusion process. Once the synthetic data have been generated, we employ  a standard decision-tree based classifier (e.g., XGBoost) for classification, as this type of model remains better suited to tabular dataset.  Experiments on a credit card fraud detection dataset demonstrate that EmDT significantly improves downstream classification performance compared to existing oversampling and generative methods, while maintaining comparable privacy protection and preserving feature correlations present in the original data. 
\end{abstract}

\section{Introduction}

With the emergence of advanced digital technology, financial fraud has become a growing concern for companies and industries, including insurance, banking, and E-commerce \cite{ali2022financial}. Recent reports indicate that the United States remains a major target for fraud, with nearly 46\% of global credit card fraud losses occurring outside its borders \cite{alrasheedi2025enhancing}. Detecting credit card fraud is crucial but challenging for financial institutions, as fraudsters quickly adopt new methods to evade detection, such as mimicking cardholder spending patterns. These trends highlight the need for reliable fraud detection systems to mitigate substantial financial risks.

Many existing fraud prediction models have been applied in identifying fraudulent behavior, such as machine learning classifiers and deep neural networks. However, their performance is often impaired by the severe class imbalance present in real-world datasets, where the proportion of fraudulent transactions is very low \cite{chen2025deep}. This class imbalance results in the model becoming biased toward the majority class and misclassifying the fraudulent samples. To combat this behavior, generating synthetic minority (fraudulent) samples has become a crucial strategy for improving model robustness.

Traditional oversampling techniques, such as SMOTE \cite{chawla2002smote}, address class imbalance by interpolating between existing fraud samples. Despite their effectiveness in balancing class distributions, they often struggle to capture the complex and multi-modal nature of real-world fraud behavior \cite{zhi2025research}. More recently, generative diffusion models have demonstrated promising generative capability in various domains, including speech processing \cite{kong2020diffwave}, computer vision \cite{song2020score}, and tabular datasets \cite{kotelnikov2023tabddpm}. Although diffusion models learn the underlying data distribution by progressively denoising the data, they may fail to preserve intricate feature relationships \cite{zhang2025temporal}. 

\begin{figure}[!t]
  \centering
  \includegraphics[width=6in]{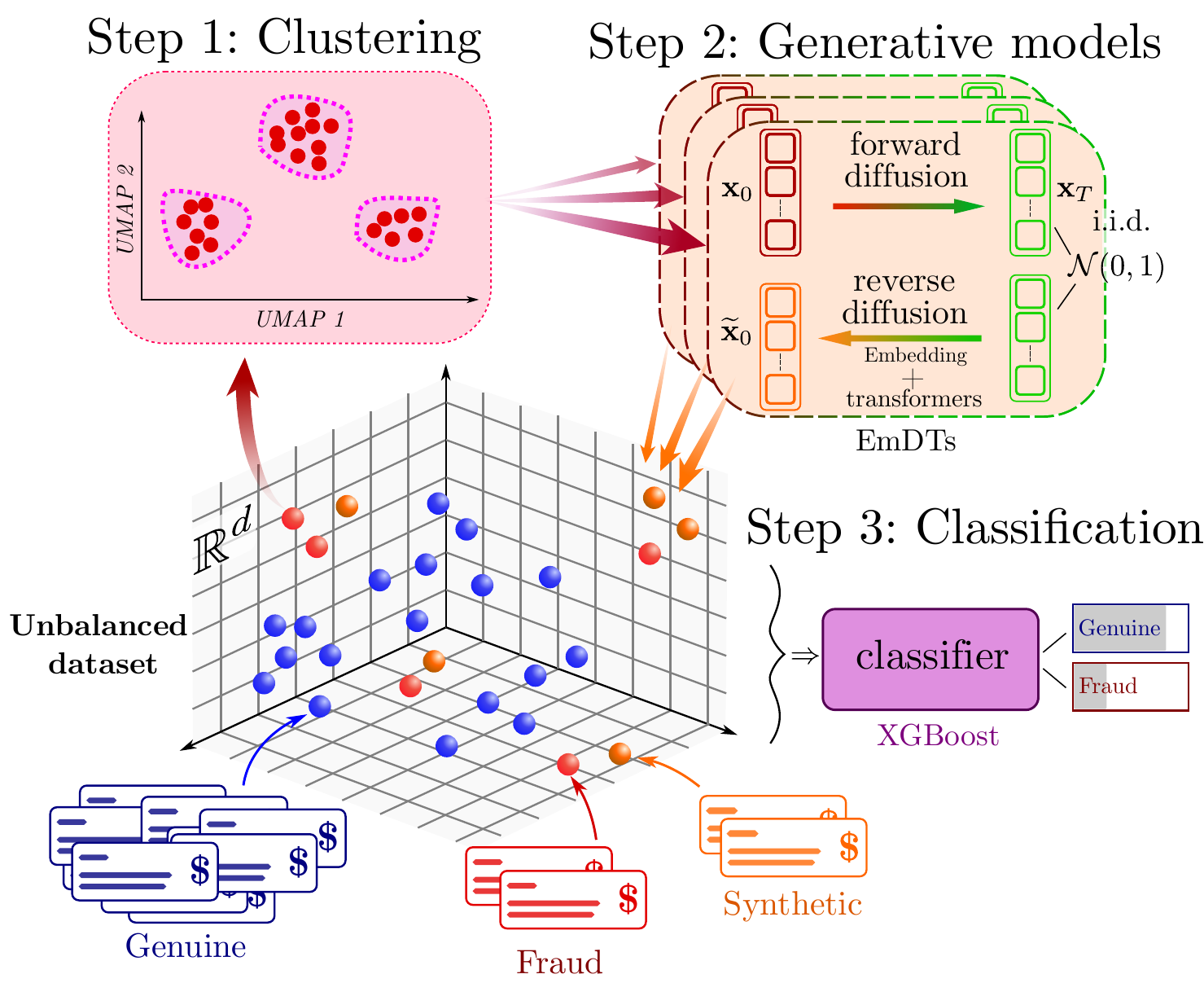}
  \caption{Overview of the proposed EmDT architecture. Starting from an imbalanced transaction dataset, the minority samples are first projected into a two-dimensional UMAP space, where distinct fraud clusters are identified (Step 1). A separate diffusion-based generative model is then trained for each cluster (Step 2), using sinusoidal embeddings and a Transformer architecture to generate synthetic fraud samples from normal samples. The synthetic data are combined with the original dataset to train a tree-based classifier such as XGBoost (Step 3), further improving the detection of fraudulent transactions.
  } 
  \label{fig:model_overview}
\end{figure}

In this study, we introduce EmDT, a novel approach for synthesizing fraudulent transactions leveraging diffusion models. As shown in Figure \ref{fig:model_overview}, EmDT first applies UMAP clustering to identify distinct fraud patterns, then employs a Transformer architecture with sinusoidal embedding as the denoising model to capture intricate feature dependencies. To the best of our knowledge, this is the first study to apply UMAP clustering with a diffusion model for financial tabular data synthesis to address class imbalance. The results demonstrate that the proposed EmDT framework consistently outperforms existing state-of-the-art generative methods, highlighting its effectiveness for realistic fraud synthesis and enhanced classification performance.

We emphasize that deep neural network architectures are used only for synthetic data generation. For the final classification, we employ a traditional machine learning approach, namely tree-based methods (e.g., XGBoost). Indeed, for tabular datasets, such methods still tend to outperform deep neural networks. Therefore, rather than attempting to surpass them with deep neural networks, our goal is to combine the strengths of both approaches: using deep neural networks to generate synthetic data and tree-based models to classify the resulting data.

\paragraph{Related Work.}
A wide range of oversampling and generative methods has been proposed to address class imbalance in fraud detection. Traditional oversampling techniques, such as SMOTE, remain widely used in this domain \cite{sundaravadivel2025optimizing}, \cite{zhu2024enhancing}. SMOTE generates synthetic samples by interpolating between minority-class instances and often improves downstream classifier performance. However, it solely relies on linear interpolation and therefore struggles to capture complex, non-linear feature structures \cite{blagus2013smote}. 

To address this limitation, more advanced generative approaches have been proposed. Several methods, such as CTGAN \cite{xu2019modeling} and CTAB-GAN \cite{zhao2021ctab}, adopt Generative Adversarial Networks (GANs) for tabular data generation. These models employ a generator to produce synthetic samples and a discriminator to distinguish real data from generated data. Although GAN-based methods can produce realistic tabular samples, they are prone to training instability and mode collapse \cite{thanh2020catastrophic}. As an alternative, Variational Autoencoders (VAEs) have also been explored for tabular data generation. TVAE \cite{xu2019modeling} extends the VAE framework to handle mixed-type tabular features. While VAEs generally provide more stable training than GANs, they tend to produce samples with lower fidelity \cite{dai2019diagnosing}.

More recently, the Denoising Diffusion Probabilistic Model (DDPM), introduced by Ho et al. \cite{ho2020denoising}, has emerged as an advanced paradigm in generative modeling. DDPMs learn the data distribution by gradually adding noise in a forward process and reversing it through iterative denoising. This process enables stable training and avoids mode collapse. Recent studies, including TabDDPM \cite{kotelnikov2023tabddpm} and FinDiff \cite{sattarov2023findiff}, demonstrate that diffusion models are well-suited for tabular data generation and can achieve a favorable balance between downstream performance and privacy. However, most existing diffusion-based methods for fraud detection employ multi-layer perceptrons (MLPs) as noise predictors. In contrast, our work employs a Transformer-based denoising architecture to more effectively capture feature dependencies in financial tabular data.
 
The structure of this paper is as follows: Section \ref{Sec:Methodology} introduces the fundamentals of diffusion models and presents the proposed methodology, including the EmDT architecture and its training process. Sections \ref{Sec:Experiment} and \ref{Sec:Results} describe the dataset, experimental setup, and experimental results. Finally, Section \ref{Sec:Conclusion} concludes the paper and discusses directions for future research.

\section{Methodology} \label{Sec:Methodology}

In this section, we provide a brief review of diffusion models and describe the proposed EmDT architecture, including its training procedure on the distinct fraud clusters.

\subsection{Diffusion Models}

\begin{figure}[h!]
  \centering
  \includegraphics[width=6in]{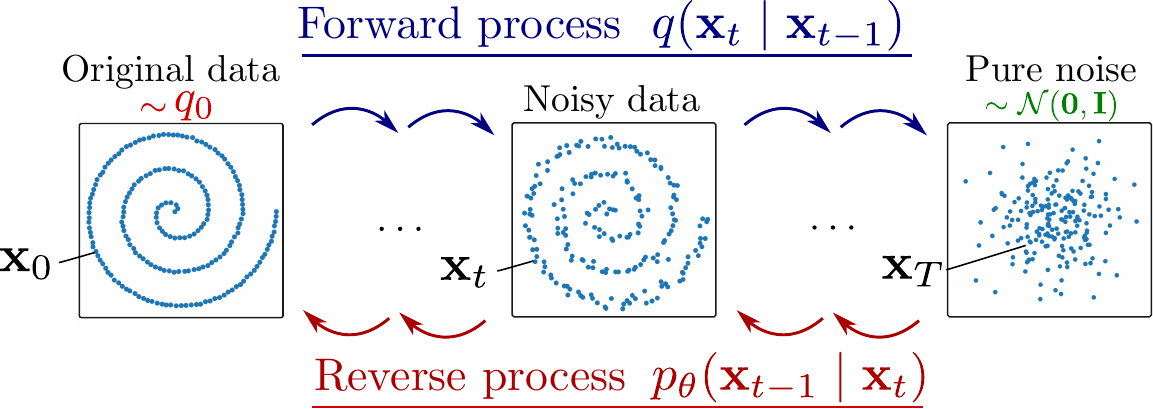}
  \caption{Illustration of forward and reverse processes in the diffusion model. The forward process $q(\mathbf{x}_t \mid \mathbf{x}_{t-1})$ progressively corrupts the original data distribution $q_0$ (left) by adding Gaussian noise to each sample ${\bf x}_0$ over timesteps, eventually transforming the data into pure Gaussian noise $\mathbf{x}_T \sim \mathcal{N}(\mathbf{0}, \mathbf{I})$ (right). The reverse process $p_\theta\left(\mathbf{x}_{t-1} \mid \mathbf{x}_{t} \right)$ learns to reverse this noising process by iteratively denoising samples, recovering the data distribution from pure noise.}
  \label{fig:ddpm-model}
\end{figure}

The Denoising Diffusion Probabilistic Model (DDPM), introduced by Ho et al. \cite{ho2020denoising}, is a stochastic diffusion process that consists of two phases: (1) a forward diffusion process, which gradually adds random noise to the data over $T$ timesteps, and (2) a reverse process, which progressively learns to denoise the data and generate new samples (see Figure~\ref{fig:ddpm-model}).

In the forward process, Gaussian noise is added incrementally to the original data $\mathbf{x}_0 \sim q_0$ over $T$ timesteps. This process can be defined as: 
\begin{equation}
q(\mathbf{x}_t \mid \mathbf{x}_{t-1}) = \mathcal{N}\left(\mathbf{x}_t; \sqrt{1 - \beta_t} \, \mathbf{x}_{t-1}, \, \beta_t \mathbf{I} \right)
\label{eq:DDPM-forward}
\end{equation}
where $\mathbf{x}_t$ is the data at the timestep $t$, and $\{\beta_t\}_{t=1}^T$ is the predefined noise schedule applied at each timestep $t$. Sampling $\mathbf{x}_t$ can also be expressed directly in terms of the original data $\mathbf{x}_0$:
\begin{equation}
\begin{aligned}
\mathbf{x}_t & =\sqrt{\bar{\alpha}_t} \, \mathbf{x}_0 + \sqrt{1-\bar{\alpha}_t} \, \boldsymbol{\epsilon}_t, \quad \boldsymbol{\epsilon}_t \sim \mathcal{N}(\mathbf{0}, \mathbf{I})
  \label{eq:DDPM-forward-x0}
  \end{aligned}
\end{equation}
where $\alpha_t := 1- \beta_t$ and $\bar{\alpha}_t := \prod_{i=1}^t\alpha_i$. As the timestep $t$ increases, this process gradually transforms the data $\mathbf{x}_t$ into a standard Gaussian distribution.

In the reverse diffusion process, the objective is to reverse the forward diffusion process and recover the original data distribution from noise. Starting from a Gaussian noise sample $\mathbf{x}_T \sim \mathcal{N}(\mathbf{0}, \mathbf{I})$, the reverse diffusion process is modeled by a neural network as follows~\cite{ho2020denoising}: 
\begin{equation}
p_\theta\left(\mathbf{x}_{t-1} \mid \mathbf{x}_{t} \right) =  \mathcal{N}\left(\mathbf{x}_{t-1}; \boldsymbol{\mu}_\theta \left(\mathbf{x}_t, t \right), \sigma^2_t \mathbf{I}\right)
  \label{eq:DDPM-Reverse-rho}
\end{equation}
where the mean $\boldsymbol{\mu}_\theta$ and variance $\sigma_t^2$ are given by:
\begin{equation}
    \boldsymbol{\mu}_\theta \left(\mathbf{x}_t, t \right)  =\frac{\sqrt{\bar{\alpha}_{t-1}} \beta_t}{1-\bar{\alpha}_t} \,\boldsymbol{\epsilon}_\theta(\mathbf{x}_t, t) + \frac{\sqrt{\alpha_t}\left(1-\bar{\alpha}_{t-1}\right)}{1-\bar{\alpha}_t} \,\mathbf{x}_t \quad \text{and} \quad \sigma_t^2 = \frac{1-\bar{\alpha}_{t-1}}{1-\bar{\alpha}_t}\beta_t.
  \label{eq:DDPM-Reverse-mu}
\end{equation}
The neural network $\boldsymbol{\epsilon}_\theta(\mathbf{x}_t, t)$ is trained to predict the Gaussian noise $\boldsymbol{\epsilon}_t \sim \mathcal{N}(\mathbf{0}, \mathbf{I})$ in \eqref{eq:DDPM-forward-x0}. Accordingly, the training objective is to minimize the corresponding mean-squared error over each timestep $t$ and each sample $\mathbf{x}_0$:
\begin{equation}
\label{eq:DDPM-loss-function}
\mathcal{L}[\theta]
= \underset{\mathbf{x}_0\sim q_0,\\\, t, \, \boldsymbol{\epsilon}_t\sim \mathcal{N}(0,{\bf I})}{\mathbb{E}} 
\left[ \, \left\| \boldsymbol{\epsilon}_t - \boldsymbol{\epsilon}_\theta(\mathbf{x}_t, t) \right\|^2 \, \right]
\end{equation}

\begin{algorithm}[H]
\caption{DDPM Training Process}
\label{alg:training_process}
\textbf{Input:} Original data $\mathbf{x}_0^{(1)},...,\mathbf{x}_0^{(n)} \in \mathbb{R}^d$, number of timesteps $T \in \mathbb{N}$, denoising model parameters $\theta$ \\
\textbf{Output:} The trained denoising model parameters $\theta^*$
\begin{algorithmic}[1]
\REPEAT
    \FOR{$i = 1$ to $N$}
        \STATE Sample $t_i\in [1,..,T]$, sample $\boldsymbol{\epsilon}_i \sim \mathcal{N}(\mathbf{0}, \mathbf{I})$
        \STATE Compute $\mathbf{x}_t^{(i)}$ from $\mathbf{x}_0^{(i)}$, $t_i$ and $\boldsymbol{\epsilon}_i$ using \eqref{eq:DDPM-forward-x0}
        \STATE Predict $\boldsymbol{\epsilon}_\theta(\mathbf{x}_t^{(i)}, t_i)$ with a neural network
    \ENDFOR
    \STATE Compute the MSE loss:
    \begin{equation}
    L(\theta) = \frac{1}{N} \sum_{i=1}^N \| \boldsymbol{\epsilon}_i - \boldsymbol{\epsilon}_\theta(\mathbf{x}_t^{(i)}, t_i) \|^2
    \label{eq:MSE-ddpm}
    \end{equation}
    \STATE Update $\theta$ using gradient descent on \eqref{eq:MSE-ddpm}
\UNTIL{converged}
\end{algorithmic}
\end{algorithm}

\begin{algorithm}[H]
\caption{DDPM Sampling Process}
\label{alg:Sampling_process}
\textbf{Input:} Number of timesteps $T \in \mathbb{N}$, the trained denoising model parameters $\theta^*$ \\
\textbf{Output:} Generated data $\tilde{\mathbf{x}} \in \mathbb{R}^d$
\begin{algorithmic}[1]
\STATE Sample $\mathbf{x}_T \sim \mathcal{N}(\mathbf{0}, \mathbf{I})$
\FOR{$t = T$ to $1$}
    \STATE Predict $\boldsymbol{\epsilon}_\theta(\mathbf{x}_t, t)$
    \STATE Compute mean $\boldsymbol{\mu}_\theta(\mathbf{x}_t, t)$ and variance $\sigma_t^2$ using \eqref{eq:DDPM-Reverse-mu}
    \STATE Sample $\mathbf{z} \sim \mathcal{N}(\mathbf{0}, \mathbf{I})$ if $t>0$, otherwise take $\mathbf{z}=0$
    \STATE Set $\mathbf{x}_{t-1} = \boldsymbol{\mu}_\theta(\mathbf{x}_t, t) + \sigma_t  \mathbf{z}$
\ENDFOR
\RETURN $\mathbf{x}_0$
\end{algorithmic}
\end{algorithm}

\subsection{Proposed Model: Embedding Diffusion Transformers (EmDT)}

The proposed EmDT framework is designed to enhance the model's ability to capture feature dependencies in highly imbalanced tabular datasets. The key insight behind EmDT is that the minority-class data often exhibits distinct clusters, as shown in Figure \ref{fig:model_overview}. This clustered structure motivates training diffusion models separately on each cluster, allowing the model to learn different fraud patterns. Experimental results further demonstrate that cluster-wise training and sampling improve downstream classification performance.

\begin{figure}[t!]
  \centering
  \includegraphics[width=\linewidth]{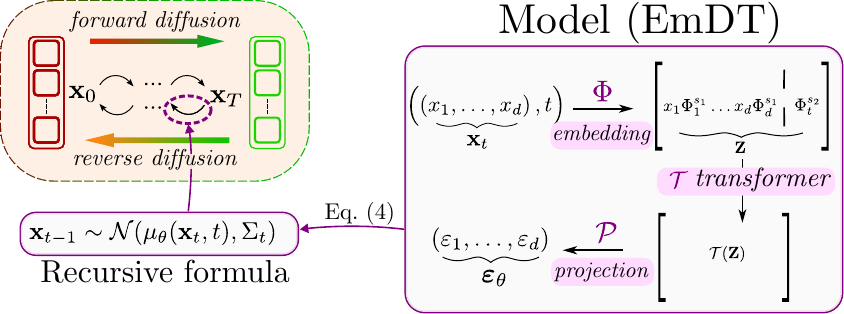}
  \caption{An Overview of the proposed \textit{EmDT} model. In the forward process, Gaussian noise is gradually added to the fraud training samples. During the reverse process, the \textit{EmDT} embeds the noisy inputs into higher-dimensional spaces and applies a Transformer to better capture feature relationships. Followed by a linear projection, the \textit{EmDT} model learns to denoise the data and generate synthetic fraud samples.}
  \label{fig:EmDT_architecture}
\end{figure}

In addition to cluster-specific training, EmDT introduces two architectural modifications within the diffusion framework to predict model Gaussian noise. First, it applies sinusoidal positional embeddings to encode feature-wise information. Second, it replaces the conventional MLP denoising network with a Transformer architecture to better capture feature dependencies. The noise prediction process can be expressed as:
\begin{equation}
    \boldsymbol{\epsilon}_\theta = \mathcal{P} \, \circ \mathcal{T} \, \circ \Phi(\mathbf{x},t)
    \label{eq: EmDT-model}
\end{equation}
where:
\begin{itemize}
    \item $\mathbf{x}$ denotes the tabular data, and $t$ is the diffusion timestep, 
    \item $\Phi$ represents sinusoidal embeddings with positional information, 
    \item $\mathcal{T}$ is a Transformer-based denoising network, and
    \item $\mathcal{P}$ is a linear projection that maps the latent representation back to the original feature space. 
\end{itemize}
This design enhances the model's ability to learn complex, multi-dimensional relationships among features. The overview of the EmDT framework, integrated with the diffusion model, is illustrated in Figure \ref{fig:EmDT_architecture}. Further details of each component are described in the following sections.

\subsubsection{Sinusoidal Positional Embedding $\Phi$}

In order to use a transformer architecture, the input vector ${\bf x}=(x_1,\dots,x_d)$ and the timestep $t$ have to be embedded into a 2D tensor. We use a sinusoidal embedding function $\phi$ that maps a scalar into a $D$-dimensional vector. Applying $\phi$ to each feature leads to the following embedding $\Phi$:
\begin{equation}
\begin{aligned}
    \Phi (\mathbf{x},t) &= [\, x_1 \, \psi_1^{(s_1)}, \, \dots, \,x_d \, \psi_d^{(s_1)}, \, \psi_t^{(s_2)} \, ] 
    \quad \text{ with} \quad \psi_j^{(s)} =
    \begin{bmatrix}
    \sin \left( s \cdot j  \right) \\
    \cos \left( s \cdot j  \right) \\
    \vdots \\
    \sin \left(\gamma^{D/2-1} s \cdot j  \right) \\
    \cos \left(\gamma^{D/2-1} s \cdot j  \right)
    \end{bmatrix},
\end{aligned}
\label{eq:Psi-embedding}
\end{equation}
where the parameter $\gamma = 10,000^{-2/D}$ controls the frequency scaling, and $s$ determines the input scaling. As a result, the embedding $\Phi$ requires three hyper-parameters to be tuned: $D$ the latent-space dimension (required to be even), $s_1$ the scaling for the feature vector ${\bf x}$, and $s_2$ the scaling for the timestep $t$. Notice that there are no trainable parameters in the embedding $\Phi$.

\subsubsection{Denoising Diffusion Model - Transformer $\mathcal{T}$}

After embedding the input features and timestep into a tensor, the EmDT model uses a transformer architecture~\cite{vaswani2017attention} to capture feature dependencies. Let $\mathbf{Z} = \Phi(\mathbf{x}, t) \in \mathbb{R}^{(d+1) \times D}$ denote the embedded feature matrix obtained from the feature vector $\mathbf{x}$ via the sinusoidal positional embeddings in~\eqref{eq:Psi-embedding}. The transformer block can be written as the composition of two transformations, each with a residual skip connection:
\begin{equation}
\begin{aligned}
    \mathbf{Z}' &= \mathbf{Z} + \texttt{Attention}(\mathbf{Z}) \\
    \mathcal{T}(\mathbf{Z}) &= \mathbf{Z}' + \texttt{FFN}(\mathbf{Z}')
\end{aligned}
    \label{eq:transformer-denoising}
\end{equation}
The \texttt{Attention} transformation computes pairwise feature dependencies by mapping $\mathbf{Z}$ into query $\mathbf{Q}$, key $\mathbf{K}$, and value $\mathbf{V}$ matrices through learnable linear projections~\cite{vaswani2017attention}:
\begin{equation}
    \mathbf{Q} = \mathbf{Z} \mathbf{W}_Q, \quad
    \mathbf{K} = \mathbf{Z} \mathbf{W}_K, \quad
    \mathbf{V} = \mathbf{Z} \mathbf{W}_V
\end{equation}
where $\mathbf{W}_Q, \mathbf{W}_K, \mathbf{W}_V \in \mathbb{R}^{D \times d_k}$ are learnable parameter matrices, and $d_k$ is the dimension of the queries and keys. The scaled dot-product attention is then computed as:
\begin{equation}
    \tilde{\mathbf{Z}} = \texttt{Attention}(\mathbf{Z})
    = \texttt{softmax} \left( \frac{\mathbf{Q} \mathbf{K}^T}{\sqrt{d_k}} \right) \mathbf{V}
\end{equation}
The feedforward transformation \texttt{FFN} is a two-layer MLP with ReLU activation, applied independently to each row of $\tilde{\mathbf{Z}}$:
\begin{equation}
    \texttt{FFN}(\tilde{\mathbf{Z}}) = \sigma(\tilde{\mathbf{Z}}\mathbf{W}_1)\mathbf{W}_2
\end{equation}
where $\sigma$ is the ReLU function applied element-wise, and $\mathbf{W}_1, \mathbf{W}_2$ are weight matrices to be learned. Note that bias vectors and layer normalization are omitted for clarity.

\medskip   

Finally, following the Transformer block, the output $\mathcal{T}(\mathbf{Z})$ is passed through an additional linear projection layer $\mathcal{P}$ to transform each row into a scalar getting the output into the same dimension as the original feature vector. In total, there are six parameter matrices to train: $\mathbf{W}_Q, \mathbf{W}_K, \mathbf{W}_V$ (attention layer), $\mathbf{W}_1, \mathbf{W}_2$ (feedfoward), $\mathbf{W}_{\mathcal{P}}$ (projection matrix).

\subsection{Training EmDT}

\begin{figure}[h!]
  \centering
  \includegraphics[width=.9\linewidth]{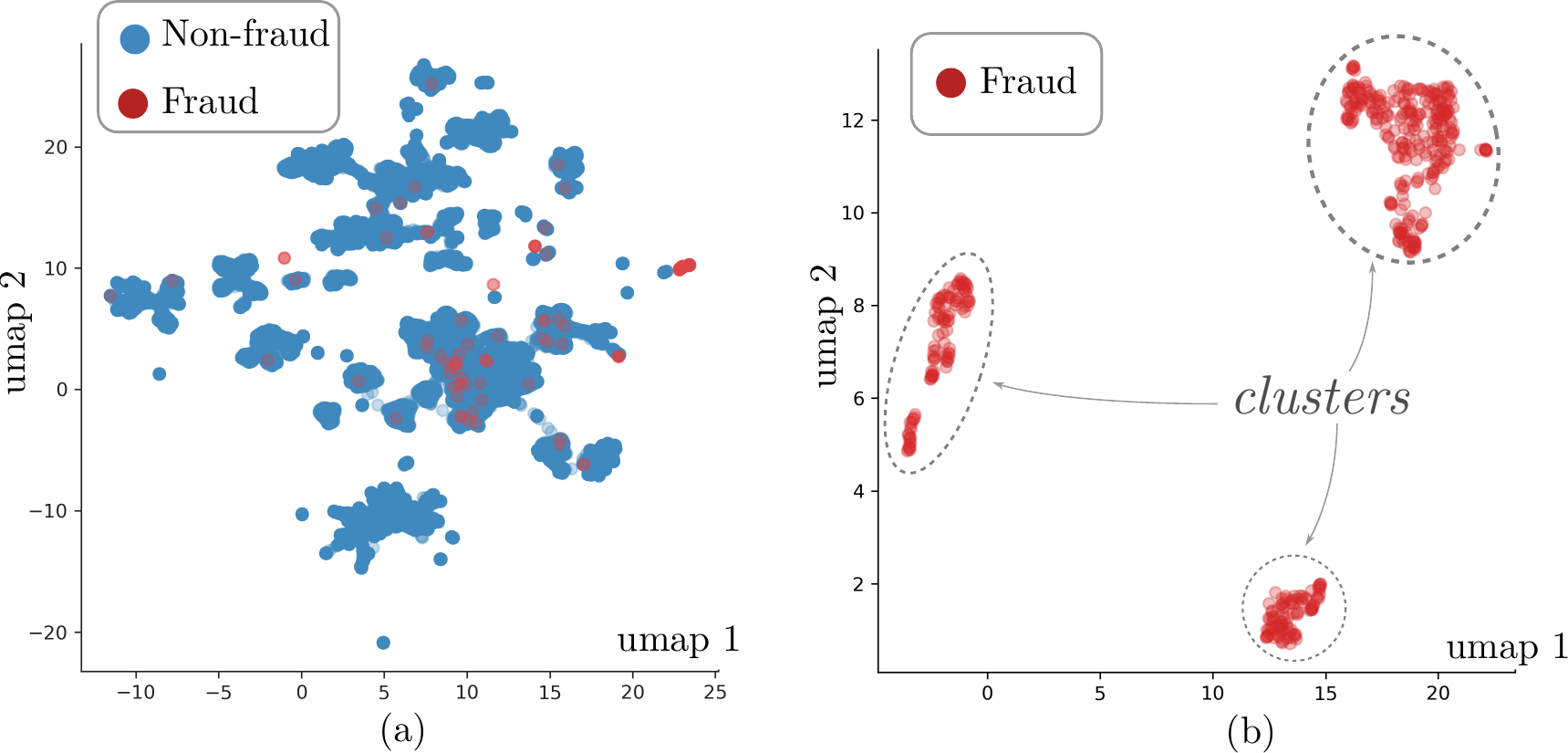}
  \caption{%
  \textbf{Left:} UMAP visualization of the Credit Card dataset ($N = 284{,}807$ samples, $d=29$ features).
  Fraudulent transactions (minority class, $n = 492$, $0.17\%$) are shown in red,
  while legitimate transactions (majority class, ${\approx}99.83\%$) are shown in blue.
  The substantial overlap between classes highlights the difficulty of the classification task.
  \textbf{Right:} UMAP projection restricted to fraudulent transactions only,
  revealing a structure of 3 distinct clusters.%
  }
  \label{fig:umap_creditcard}
\end{figure}

When the fraudulent samples are embedded into a 2D space using \texttt{UMAP}, we observe three clearly separated clusters, as shown in Figure~\ref{fig:umap_creditcard}. To leverage this structure, we train a separate DDPM model for each cluster, using the proposed embedding function and the Transformer-based denoising network. This strategy allows the model to capture the distinct distributions within the fraudulent data and generate synthetic samples that may potentially enhance model performance. The overall training and sampling procedure of EmDT is summarized in Algorithm \ref{algo:EmDT-training}.

\begin{algorithm}[H]
\caption{EmDT Training and Sampling Procedure}
\label{algo:EmDT-training}
\textbf{Input:} Fraud data $\mathbf{X}_{\text{fraud}}=[{\bf x}^{(1)},..,{\bf x}^{(n)}]$ with $ {\bf x}^{(i)}\in \mathbb{R}^d$, total number of synthetic samples $m$ \\
\textbf{Output:} Synthetic fraud samples $\tilde{\mathbf{x}}^{(1)},..,\tilde{\mathbf{x}}^{(m)} \in \mathbb{R}^d$ 
\begin{algorithmic}[1]
    \STATE Embed fraud data in 2D using UMAP:\\ \centerline{$\mathbf{U}_{\text{fraud}} \gets \texttt{UMAP}(\mathbf{X}_{\text{fraud}})$}
    \STATE Cluster $\mathbf{U}_{\text{fraud}}$ using KMeans into 3 groups: \\ \centerline{$\mathbf{X}_{\text{fraud}}^{(k)} \gets \texttt{KMeans}(\mathbf{U}_{\text{fraud}}, 3),\quad k=1,2,3 $}
    \FOR{$k = 1$ to $3$}
        \STATE Train a DDPM model $\mathcal{D}_k$ on $\mathbf{X}_{\text{fraud}}^{(k)}$, using \eqref{eq: EmDT-model} and Algorithm \ref{alg:training_process}
        \STATE Generate $m_k$ synthetic samples using $\mathcal{D}_k$ and Algorithm \ref{alg:Sampling_process}, where $m_k = \text{round}\!\left(m \cdot \frac{|\mathbf{X}_{\text{fraud}}^{(k)}|}{n}\right)$
    \ENDFOR
    \STATE Append all synthetic samples as $\tilde{\mathbf{x}}^{(1)},..,\tilde{\mathbf{x}}^{(m)}$
\end{algorithmic}
\end{algorithm}

The EmDT model was implemented in \texttt{PyTorch} \cite{paszke2019pytorch}. Hyperparameters for both the Transformer and diffusion models were selected through Optuna search \cite{akiba2019optuna}. Table \ref{tab:Hyperparameter} summarizes the search ranges and selected values. The model is trained for $150$ epochs using the Adam optimizer \cite{adam2014method} to minimize the loss described in \eqref{eq:MSE-ddpm}.  We set the diffusion process to $1000$ timesteps, and the noise variance $\beta$ followed a linear schedule from $0.001$ to $0.02$. In the Transformer block, we use two attention heads and set the feedforward dimension to 128.

\renewcommand{\arraystretch}{1.2}
\begin{table}[h!]
\centering
\caption{Main hyperparameters for EmDT.}
\label{tab:Hyperparameter}
\begin{tabular}{>{\hspace{0em}}l >{\hspace{0em}}c >{\hspace{0em}}l} 
\toprule
\textbf{Hyperparameter} & \textbf{Notation} & \textbf{Value/Search Space} \\
\hline
Feature dimension & $d$  & 29 \\
\midrule
Embedding dimension & $D$  & \{32, 64, 128\} \\
Feature embedding scale & $s_1$  & \{1, 10, 50, 100, 500\} \\
Time embedding scale &$s_2$  & \{0.5, 1, 2\} \\
\midrule
Diffusion timesteps  & $T$  & 1,000 \\
Learning rate & - & [0.00001, 0.003] \\
Batch size & -  & \{64, 128, 256\} \\
Multi-head & -  & 2 \\
Feed forward dimension & - & same as Embedding dimension \\
\bottomrule
\end{tabular}
\end{table}

\begin{figure}[h!]
  \centering
  \includegraphics[width=.7\linewidth]{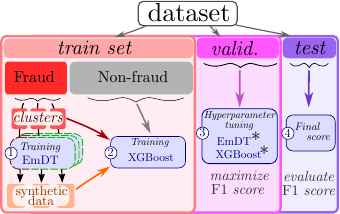}
  \caption{Overview of the performance evaluation workflow for generative models. The procedure is divided into four steps. First, generative models based on EmDT are trained. Second, a tree-based classifier (XGBoost) is optimized using the training set and the synthetic data. Third, the hyperparameters of the EmDT and classifier are optimized based on the F1 score on the validation set. Fourth, the classifier is evaluated on the test set.}
  \label{fig:Workflow-ddpm}
\end{figure}

The dataset was split into a 60\% training set, a 20\% validation set, and a 20\% test set. Following prior work \cite{roy2024frauddiffuse}, we train the generative model on fraudulent samples and generate synthetic fraud instances to double the size of the minority class. The validation set was used to tune the hyperparameters based on F1-score, and the final performance was reported on the held-out test set. An overview of the evaluation workflow is illustrated in Figure \ref{fig:Workflow-ddpm}. 

\section{Experiment} 
\label{Sec:Experiment}

We perform the experiments on an NVIDIA RTX A4000 GPU with 16 GB memory. To ensure robustness, all reported results are averaged over 10 runs with different random seeds.

\subsection{Dataset}
The dataset used in our experiments is the publicly available and widely referenced credit card fraud detection dataset from Kaggle \cite{dal2015calibrating}. A visualization and summary of the dataset is provided in figure \ref{fig:umap_creditcard}. It contains credit card transactions made by European cardholders over the span of two-day period in September $2013$. The dataset includes $31$ features, $28$ of which are principal components transformed using Principal Component Analysis (PCA), while the remaining two are \textit{Time}, which specifies the seconds elapsed between each transaction and the first transaction in the dataset, and \textit{Amount}, which is the transaction amount. The variable 'Class' contains the target labels, with a value of $1$ for fraudulent transactions and $0$ otherwise.

Notably, this dataset exhibits a severe class imbalance, which has only $492$ frauds out of $284,807$ total transactions, i.e., the positive class (frauds) comprises $0.172\%$ of all transactions. Such an imbalance poses a major challenge for model generalization and fraud detection accuracy, as fraudulent samples are heavily outnumbered by legitimate ones.

Furthermore, in fraud detection, obtaining publicly available datasets is inherently difficult due to privacy concerns. Datasets that include personally identifiable information must be anonymized or removed before release. The dataset we use is therefore unique because it is the only publicly accessible dataset of credit card transactions that reflects real-world usage patterns, rather than simulated data.

Figure~\ref{fig:umap_creditcard}(a) presents a visualization of the credit card dataset. In this dataset, fraudulent transactions and legitimate transactions are closely clustered, making the two classes difficult to distinguish. This overlap poses challenges for generating realistic fraudulent samples. However, when we focus only on fraudulent transactions, as shown in Figure~\ref{fig:umap_creditcard}(b), the fraudulent data further separates into three clearly defined clusters. This observation indicates that the fraud class contains multiple distinct patterns. Such structure can be leveraged to generate more representative synthetic fraud samples and may improve downstream classification performance. To the best of our knowledge, this clustering behavior has not been reported in prior studies.

During preprocessing, the feature \textit{Time} was excluded because it does not represent meaningful temporal information and may lead to overfitting \cite{leevy2023comparative}. The feature \textit{Amount} was normalized, and no missing values were observed in the dataset.

\subsection{Baselines}

We compare the proposed EmDT with four state-of-the-art methods for synthetic tabular data generation. These methods include a classical statistical oversampling approach, SMOTE \cite{chawla2002smote}, which generates minority samples by interpolating between nearby minority points. We also consider three deep learning-based generative models: TVAE \cite{xu2019modeling}, CTGAN \cite{xu2019modeling}, and TabDDPM \cite{kotelnikov2023tabddpm}. 

SMOTE is implemented using \texttt{Imbalanced-learn} library \cite{lemaavztre2017imbalanced} in Python. For TVAE and CTGAN, we adopt the official implementations provided in the \texttt{SDV} GitHub repository \cite{patki2016synthetic}. TabDDPM is implemented using the authors’ official code with default hyperparameter settings.

\subsection{Evaluation Metrics}
We assess the quality of the synthetic data based on the downstream classification performance. After training each generative model, the generated samples are combined with the real training data to train an XGBoost classifier \cite{chen2016xgboost}. The model performance is then assessed on the test set using F1-score, recall, precision, and balanced accuracy (Bal-Acc).

In addition, we examine the privacy risk of the synthetic data to ensure the generative models learn the underlying data distribution rather than replicating individual training samples. Specifically, we compute the Distance to Closest Record (DCR) score \cite{platzer2021holdout}, which represents the probability that synthetic records are closer to the training set than to the test set.

\section{Results}
\label{Sec:Results}
In this section, we present the downstream classifier performance of EmDT and compare it with state-of-the-art oversampling and generative methods. In addition, we evaluate the effectiveness of the proposed model through an ablation study, a qualitative assessment of synthetic data, and a hyperparameter sensitivity analysis.

\subsection{Classification Performance Comparison}
To assess the quality of synthetic data in fraud detection, we train each generative model to augment the minority class and evaluate its effectiveness in downstream classification performance. This evaluation measures how effectively the generated samples alleviate class imbalance and facilitate model learning. The averaged results in Table \ref{tab:eval-xgboost} show that EmDT achieves the best overall performance among all competing oversampling and generative methods, and also outperforms training on the original dataset without augmentation. In particular, EmDT attains the highest F1-score, recall, precision, and balanced accuracy (Bal-Acc). These results indicate that the synthetic samples generated by EmDT provide useful representations of fraudulent transactions, thereby improving classifier generalization.

We also observe that the traditional oversampling method SMOTE achieves better classification performance than deep learning-based generative models, including CTGAN, TVAE, and TabDDPM. This result suggests that deep neural generative models may be susceptible to mode collapse or unstable training, resulting in lower-quality synthetic samples that limit the performance of downstream classifiers.

Importantly, improvements in classification performance should be considered together with privacy risks. Prior work has shown that gains in predictive accuracy often come at the cost of increased privacy leakage \cite{villaizan2025diffusion}. In our experiments, SMOTE obtains the highest DCR score, indicating that its synthetic samples are closer to the training data than to the test data. This behavior is expected, as SMOTE generates samples by interpolating directly between existing data points, instead of learning the underlying distribution. In contrast, EmDT achieves a DCR score comparable to other generative models while delivering the strongest classification performance. This result indicates that EmDT effectively captures the minority-class distribution and generates high-quality synthetic samples without significantly increasing the privacy risk.

\begin{table}[H]
    \centering
    \caption{Evaluation of ML efficiency using the XGBoost classifier and DCR score. ML efficiency performance metrics include F1-score, recall, precision, and balanced accuracy (Bal-Acc). Upward arrows ($\uparrow$) indicate that higher values correspond to better performance. The privacy metric includes DCR score, which represents the probability that synthetic data are more similar to the training set than to the test set. A value close to $50\%$ indicates a good balance between resemblance to the training and test distributions. The \textit{Original} row represents the average performance on real data without data augmentation.}
    
    \label{tab:eval-xgboost}
    \begin{tabular}{l | c c c c | c}
        \toprule
        \textbf{Method} & \multicolumn{4}{c|}{\textbf{ML Efficiency}} & \textbf{Privacy}\\
        & \textbf{F1-Score $\uparrow$} & \textbf{Recall $\uparrow$} & \textbf{Precision $\uparrow$} & \textbf{Bal-Acc $\uparrow$} & \textbf{DCR} ($\approx 0.500$)  \\
        \midrule
        Original  & 0.800 \scriptsize{$\pm0.035$} & 0.743 \scriptsize{$\pm0.041$} & 0.868 \scriptsize{$\pm0.039$} & 0.871 \scriptsize{$\pm0.02$} & - \\
        \midrule
        SMOTE & 0.834 \scriptsize{$\pm0.025$} & 0.784 \scriptsize{$\pm0.031$} & 0.891\scriptsize{$\pm0.031$} & 0.892 \scriptsize{$\pm0.016$} & 0.686 \scriptsize{$\pm0.01$} \\
        CTGAN & 0.805 \scriptsize{$\pm0.033$} & 0.751 \scriptsize{$\pm0.042$} & 0.873 \scriptsize{$\pm0.074$} & 0.875 \scriptsize{$\pm0.021$} & \textbf{0.526} \scriptsize{$\pm0.06$} \\
        TVAE & 0.820 \scriptsize{$\pm0.01$} & 0.778 \scriptsize{$\pm0.019$} & 0.869 \scriptsize{$\pm0.035$} & 0.889 \scriptsize{$\pm0.009$} & \textbf{0.528} \scriptsize{$\pm0.07$} \\
        TabDDPM  & 0.816 \scriptsize{$\pm0.019$} & 0.767 \scriptsize{$\pm0.032$} & 0.873 \scriptsize{$\pm0.036$} & 0.884 \scriptsize{$\pm0.016$} & 0.578 \scriptsize{$\pm0.04$} \\
        \textbf{EmDT (ours)}   & \textbf{0.849} \scriptsize{$\pm0.021$} & \textbf{0.791} \scriptsize{$\pm0.025$} & \textbf{0.916} \scriptsize{$\pm0.025$} & \textbf{0.895} \scriptsize{$\pm0.012$} & 0.555 \scriptsize{$\pm0.06$} \\
        \bottomrule
    \end{tabular}
\end{table}

\begin{figure}[h!]
  \centering
  \includegraphics[width=6in]{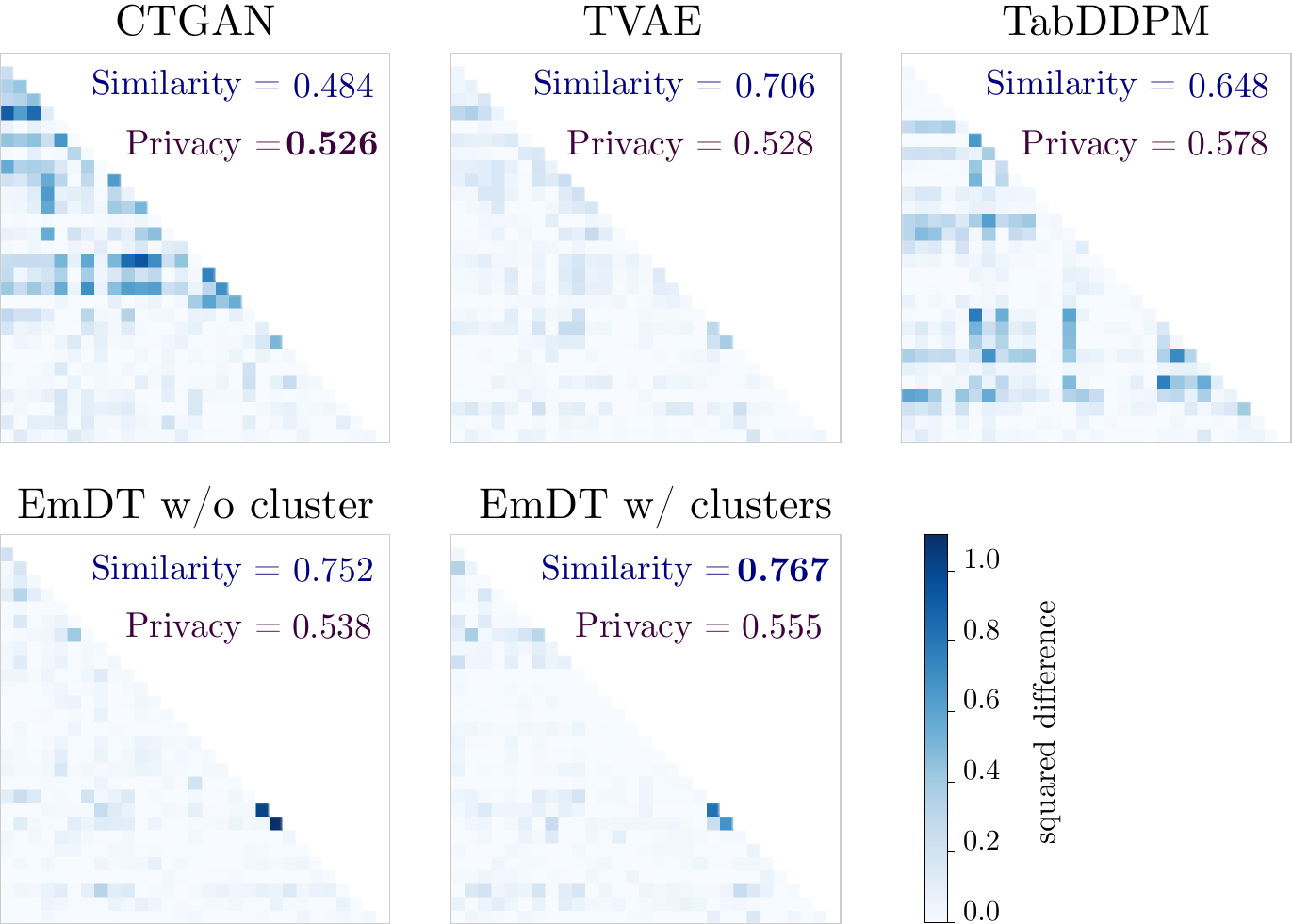}
  \caption{L2 distance between correlation matrices computed from the real and synthetic data. More intense colors indicate larger differences in correlation values. Similarity is quantified using the normalized Frobenius norm, where higher values indicate better preservation of correlation structure.}
  \label{fig:l2-corr-matrices}
\end{figure}

\begin{figure}[h!]
  \centering
  \includegraphics[width=0.9\linewidth]{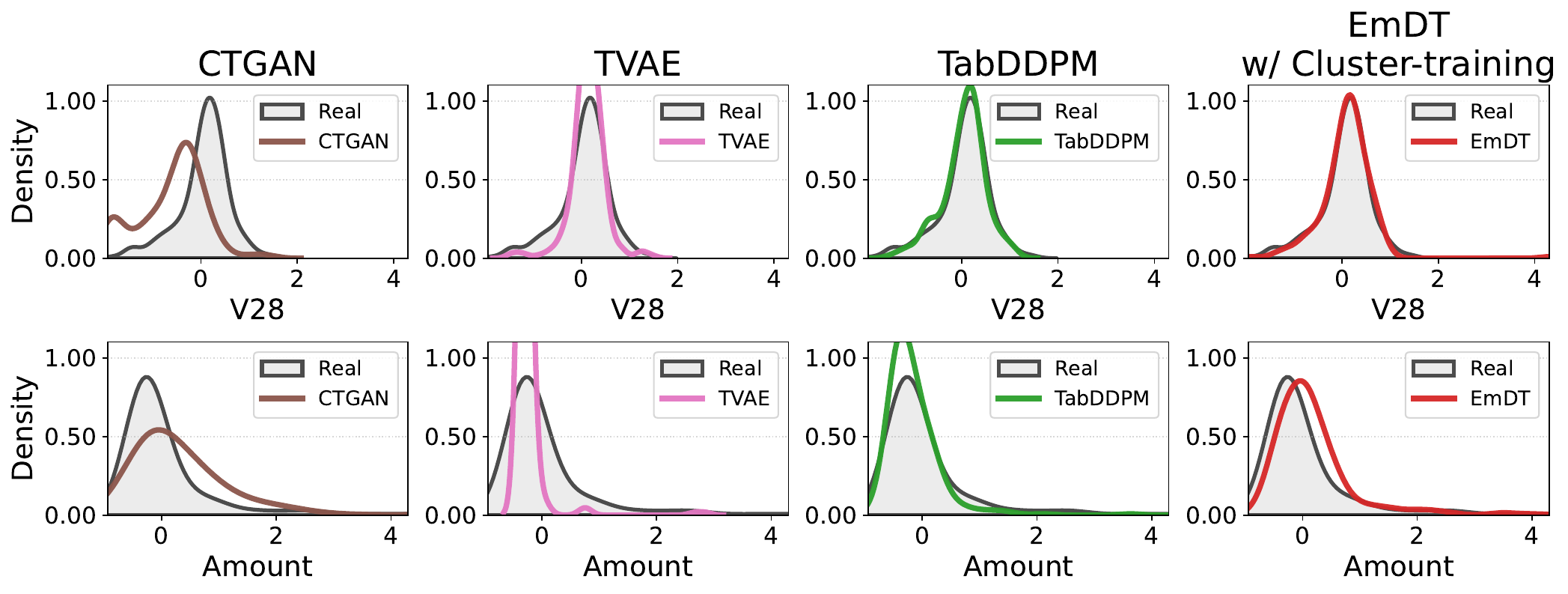}
  \caption{Comparison of feature distributions between the real dataset and the synthetic data generated by different methods. The black curves represent the distributions of real data, and the colored curves indicate the distributions of synthetic data.  }
  \label{fig:feature-density}
\end{figure}

In addition to classification performance, we provide a qualitative analysis of the synthetic data generated by CTGAN, TVAE, TabDDPM, and EmDT, as shown in Figures \ref{fig:l2-corr-matrices} and \ref{fig:feature-density}. In Figure \ref{fig:l2-corr-matrices}, we compare the L2 correlation matrix difference between the real and synthetic data, with a more intense blue color indicating a higher difference. We further quantify these differences using normalized Frobenius-norm similarity, where higher values indicate better correlation preservation. The results show that EmDT achieves the smallest correlation differences and the highest similarity among all generative models. This indicates that EmDT better captures complex feature dependencies, which is consistent with its superior downstream classification performance. In addition, EmDT with cluster-training achieves lower correlation differences than EmDT without cluster-training, highlighting the benefit of incorporating a cluster structure during training.

Figure \ref{fig:feature-density} compares the marginal distributions of two representative features, \textit{V28} and \textit{Amount}, between the real and synthetic data generated by different methods. These features are chosen because they exhibit distinct distributions: feature \textit{V28} follows a standard normal distribution, while feature \textit{Amount} has a right-skewed distribution with long tails. The distributions generated by CTGAN, TVAE, and TabDDPM exhibit noticeable deviations from the real data distribution, indicating that they may suffer from mode collapses and validating their suboptimal classification performance and the L2 distance in correlation matrices. In contrast, EmDT closely matches the real data distribution and outperforms other deep network models, demonstrating its ability to generate effective synthetic data that can improve downstream classification tasks.

\subsection{Ablation Study}
We conduct an ablation study to assess the contribution of UMAP-based cluster training to model performance. As shown in Table \ref{tab:ablation-study}, EmDT with cluster-specific training achieves comparable privacy protection while demonstrating superior classification performance in terms of F1-score, recall, precision, and balanced accuracy, compared with the model trained on the full fraud set without cluster-training. This indicates that learning from distinct clusters improves the model's predictive capability. We also observe that both versions of EmDT— with and without cluster training— outperform the original data by more than 2\%. This result highlights the importance of data augmentation for highly imbalanced fraud datasets and further supports the effectiveness of the proposed EmDT model.

\begin{table}[!t]
    \caption{Ablation study of EmDT on classification performance and privacy assessment.}
    \label{tab:ablation-study}
    \centering
    \begin{tabular}{l c c c c | c}
        \toprule
        & \multicolumn{4}{c|}{\textbf{ML Efficiency}} & \textbf{Privacy}\\
        \cmidrule{2-6}
        \textbf{Method} & \textbf{F1-Score $\uparrow$} & \textbf{Recall $\uparrow$} & \textbf{Precision $\uparrow$} & \textbf{Bal-Acc $\uparrow$} & \textbf{DCR} ($\approx 0.500$)  \\
        \midrule
        Original  & 0.800 \scriptsize{$\pm0.035$} & 0.743 \scriptsize{$\pm0.041$} & 0.868 \scriptsize{$\pm0.039$} & 0.871 \scriptsize{$\pm0.02$} & - \\
        \midrule
        EmDT w/o Cluster-training & 0.829 \scriptsize{$\pm0.027$} & 0.780 \scriptsize{$\pm0.028$} & 0.885 \scriptsize{$\pm0.034$} & 0.890 \scriptsize{$\pm0.014$} & \textbf{0.538} \scriptsize{$\pm0.04$} \\
        EmDT w/ Cluster-training  & \textbf{0.849} \scriptsize{$\pm0.021$} & \textbf{0.791} \scriptsize{$\pm0.025$} & \textbf{0.916} \scriptsize{$\pm0.025$} & \textbf{0.895} \scriptsize{$\pm0.012$} & 0.555 \scriptsize{$\pm0.06$}  \\
        \bottomrule
    \end{tabular}
\end{table}

\subsection{Hyperparameters Sensitivity Analysis}
\begin{figure}[h!]
    \centering
    % ---------- Row 1: 3 plots ----------
    \begin{subfigure}{0.32\linewidth}
        \centering
        \includegraphics[width=\linewidth]{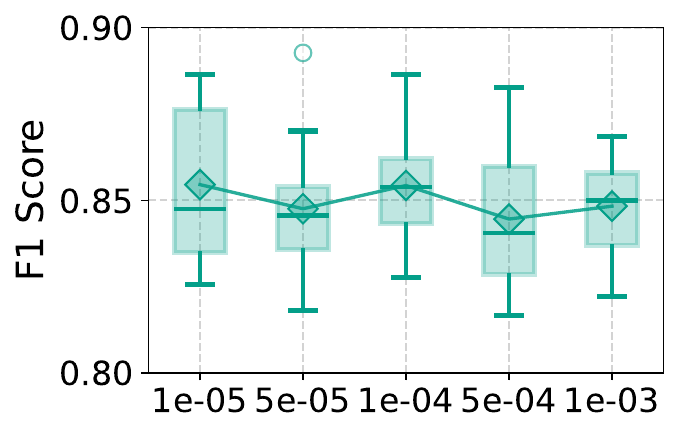}
        \caption{Learning rate}
        \label{fig:lr}
    \end{subfigure}\hspace{0.05\linewidth} % small horizontal gap
    \begin{subfigure}{0.32\linewidth}
        \centering
        \includegraphics[width=\linewidth]{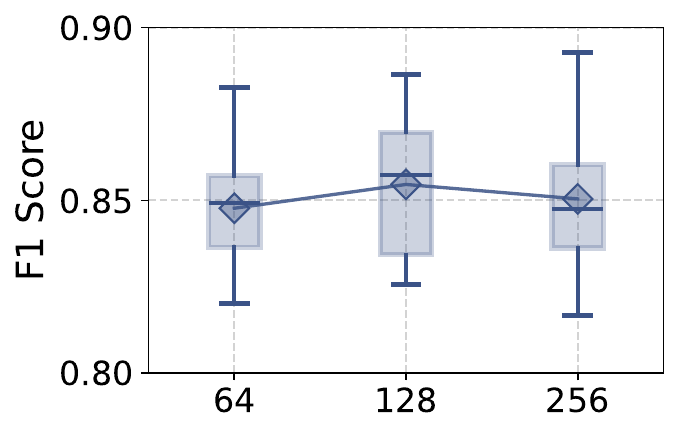}
        \caption{Batch size}
        \label{fig:batch-size}
    \end{subfigure}
    \vspace{6pt} % space between rows
    % ---------- Row 2: 2 centered plots (same width as above) ----------
    \begin{subfigure}{0.32\linewidth}
        \centering
        \includegraphics[width=\linewidth]{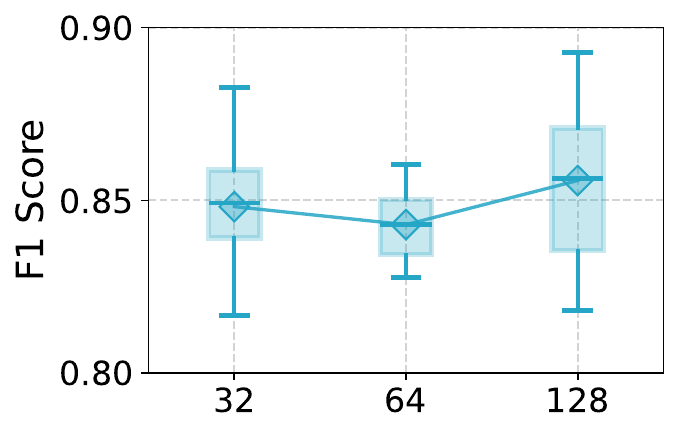}
        \caption{Embedding dimension ($D$)}
        \label{fig:emb}
    \end{subfigure}\hfill
    \begin{subfigure}{0.32\linewidth}
        \centering
        \includegraphics[width=\linewidth]{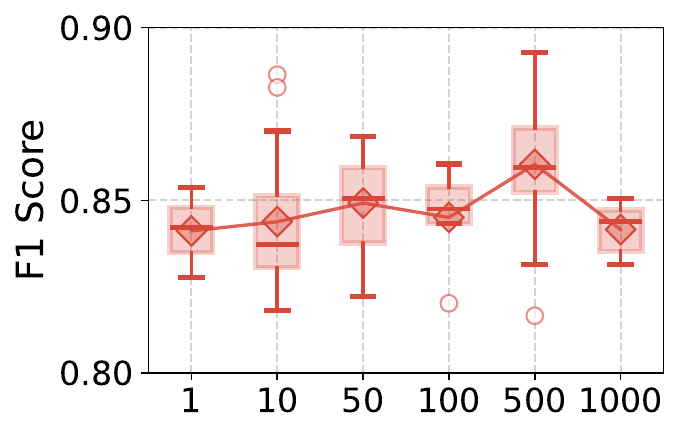}
        \caption{Feature embedding scale ($s_1$)}
        \label{fig:scale-x}
    \end{subfigure}\hfill
    \begin{subfigure}{0.32\linewidth}
        \centering
        \includegraphics[width=\linewidth]{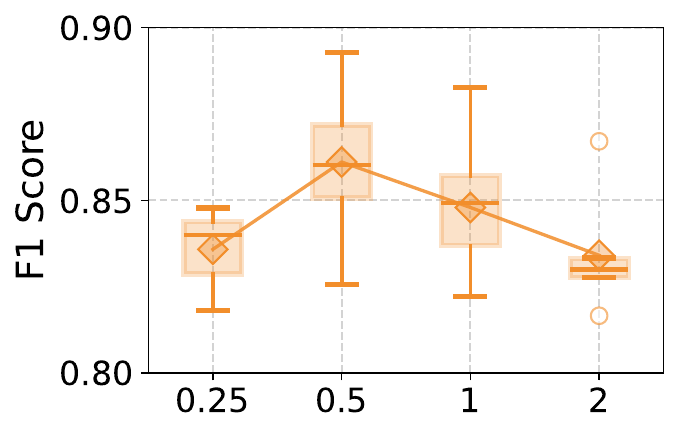}
        \caption{Time embedding scale ($s_2$)}
        \label{fig:scale-t}
    \end{subfigure}
    \caption{Effect of hyperparameter settings on EmDT performance. The boxplots show the distribution of F1-scores for each hyperparameter value, with diamond markers indicating the average performance.}
    \label{fig:hparam_sensitivity}
\end{figure}

In this section, we examine the impact of key hyperparameters on the F1-score performance of the EmDT model. We vary five hyperparameters: the learning rate, batch size, embedding dimension ($D$), feature embedding scale ($s_1$), and time embedding scale ($s_2$). As shown in Figure \ref{fig:hparam_sensitivity}, EmDT exhibits stable performance across a wide range of learning rates, batch sizes, and embedding dimensions, suggesting that these parameters have limited influence on model performance. In contrast, the embedding scales have a more noticeable effect. The model performance peaks at a feature embedding scale of $500$ and a time embedding scale of $0.5$. This behavior can be explained by the relative magnitudes of the inputs. After preprocessing, each feature approximately follows a standard normal distribution, with magnitude on the order of $10^0$, whereas the diffusion timestep is sampled from a uniform distribution $\mathcal{U}([0,1000])$, corresponding to a scale order of $10^3$. With the selected scaling factors, the feature embeddings span approximately $[0,500]$, while the timestep embeddings have a standard deviation of $500$. This scaling places both embeddings on comparable ranges, ensuring a balanced contribution during the denoising process.

Overall, EmDT demonstrates robust performance across different hyperparameter settings, with fluctuations within approximately $\pm1\%$. Although the best performance is observed with an embedding dimension of $128$, a feature embedding scale of $500$, and a time embedding scale of $0.5$, extensive hyperparameter tuning is not necessary to obtain competitive results.

\section{Conclusion and Future Work} 
\label{Sec:Conclusion}

In this study, we proposed EmDT, a cluster-guided diffusion model for synthetic data generation in fraud detection.  This framework is designed to address severe class imbalance by incorporating sinusoidal positional embedding, UMAP-based cluster training, and a transformer-based diffusion model. With these introductions, EmDT generates realistic fraudulent samples that align closely with minority-class distribution while preserving feature density and correlation structures. We have demonstrated that EmDT achieves superior performance compared with state-of-the-art oversampling and generative methods, including SMOTE, CTGAN, TVAE, and TabDDPM. At the same time, the model maintains competitive results in privacy evaluation and statistical similarity, demonstrating its ability to balance utility and privacy.

Despite these promising results, the generalizability of EmDT to other domains - such as healthcare diagnostics, insurance fraud detection, and network intrusion detection - has not yet been fully explored. Extending the model to additional datasets would provide a more comprehensive understanding of its robustness and limitations.

Another possible extension is inspired by Perpendicular Negative Prompting
(Perp-Neg) \cite{armandpour2023re}, a data sampling strategy that leverages both majority and minority class distributions to generate minority samples that remain faithful to minority patterns without drifting toward the majority distribution. Such an idea could be integrated into the diffusion process to further control data generation. Additionally, we can combine our cluster-based training strategy with the Gaussian Copula models \cite{patki2016synthetic}, which have demonstrated effectiveness in domains such as credit risk modeling \cite{hochrainer2018integrating}, signal processing \cite{morteza2022novel}, and climate science \cite{tedesco2023gaussian}. This combination could strengthen EmDT by preserving feature relationships and privacy.

\clearpage
\bibliographystyle{plain}
\bibliography{references}

@article{adam2014method,
  title={A method for stochastic optimization},
  author={Adam, Kingma DP Ba J and others},
  journal={arXiv preprint arXiv:1412.6980},
  volume={1412},
  number={6},
  year={2014}
}

@inproceedings{akiba2019optuna,
  title={Optuna: A next-generation hyperparameter optimization framework},
  author={Akiba, Takuya and Sano, Shotaro and Yanase, Toshihiko and Ohta, Takeru and Koyama, Masanori},
  booktitle={Proceedings of the 25th ACM SIGKDD international conference on knowledge discovery \& data mining},
  pages={2623--2631},
  year={2019}
}

@article{ali2022financial,
  title={Financial fraud detection based on machine learning: a systematic literature review},
  author={Ali, Abdulalem and Abd Razak, Shukor and Othman, Siti Hajar and Eisa, Taiseer Abdalla Elfadil and Al-Dhaqm, Arafat and Nasser, Maged and Elhassan, Tusneem and Elshafie, Hashim and Saif, Abdu},
  journal={Applied Sciences},
  volume={12},
  number={19},
  pages={9637},
  year={2022},
  publisher={MDPI}
}

@article{alrasheedi2025enhancing,
  title={Enhancing Fraud Detection in Credit Card Transactions: A Comparative Study of Machine Learning Models},
  author={Alrasheedi, Masad A},
  journal={Computational Economics},
  pages={1--27},
  year={2025},
  publisher={Springer}
}

@article{armandpour2023re,
  title={Re-imagine the negative prompt algorithm: Transform 2d diffusion into 3d, alleviate janus problem and beyond},
  author={Armandpour, Mohammadreza and Sadeghian, Ali and Zheng, Huangjie and Sadeghian, Amir and Zhou, Mingyuan},
  journal={arXiv preprint arXiv:2304.04968},
  year={2023}
}

@article{blagus2013smote,
  title={SMOTE for high-dimensional class-imbalanced data},
  author={Blagus, Rok and Lusa, Lara},
  journal={BMC bioinformatics},
  volume={14},
  number={1},
  pages={106},
  year={2013},
  publisher={Springer}
}

@article{chawla2002smote,
  title={SMOTE: synthetic minority over-sampling technique},
  author={Chawla, Nitesh V and Bowyer, Kevin W and Hall, Lawrence O and Kegelmeyer, W Philip},
  journal={Journal of artificial intelligence research},
  volume={16},
  pages={321--357},
  year={2002}
}

@inproceedings{chen2016xgboost,
  title={Xgboost: A scalable tree boosting system},
  author={Chen, Tianqi and Guestrin, Carlos},
  booktitle={Proceedings of the 22nd acm sigkdd international conference on knowledge discovery and data mining},
  pages={785--794},
  year={2016}
}

@article{chen2025deep,
  title={Deep learning in financial fraud detection: Innovations, challenges, and applications},
  author={Chen, Yisong and Zhao, Chuqing and Xu, Yixin and Nie, Chuanhao and Zhang, Yixin},
  journal={Data Science and Management},
  year={2025},
  publisher={Elsevier}
}

@article{dai2019diagnosing,
  title={Diagnosing and enhancing VAE models},
  author={Dai, Bin and Wipf, David},
  journal={arXiv preprint arXiv:1903.05789},
  year={2019}
}

@inproceedings{dal2015calibrating,
	author = {Dal Pozzolo, Andrea and Caelen, Olivier and Johnson, Reid A and Bontempi, Gianluca},
	booktitle = {2015 IEEE Symposium Series on Computational Intelligence},
	organization = {IEEE},
	pages = {159--166},
	title = {Calibrating probability with undersampling for unbalanced classification},
	year = {2015}
	}

@article{ho2020denoising,
  title={Denoising diffusion probabilistic models},
  author={Ho, Jonathan and Jain, Ajay and Abbeel, Pieter},
  journal={Advances in neural information processing systems},
  volume={33},
  pages={6840--6851},
  year={2020}
}

@article{hochrainer2018integrating,
  title={Integrating systemic risk and risk analysis using copulas},
  author={Hochrainer-Stigler, Stefan and Pflug, Georg and Dieckmann, Ulf and Rovenskaya, Elena and Thurner, Stefan and Poledna, Sebastian and Boza, Gergely and Linnerooth-Bayer, Joanne and Br{\"a}nnstr{\"o}m, {\AA}ke},
  journal={International Journal of Disaster Risk Science},
  volume={9},
  number={4},
  pages={561--567},
  year={2018},
  publisher={Springer}
}

@article{kong2020diffwave,
  title={Diffwave: A versatile diffusion model for audio synthesis},
  author={Kong, Zhifeng and Ping, Wei and Huang, Jiaji and Zhao, Kexin and Catanzaro, Bryan},
  journal={arXiv preprint arXiv:2009.09761},
  year={2020}
}

@inproceedings{kotelnikov2023tabddpm,
  title={Tabddpm: Modelling tabular data with diffusion models},
  author={Kotelnikov, Akim and Baranchuk, Dmitry and Rubachev, Ivan and Babenko, Artem},
  booktitle={International Conference on Machine Learning},
  pages={17564--17579},
  year={2023},
  organization={PMLR}
}

@article{leevy2023comparative,
  title={Comparative analysis of binary and one-class classification techniques for credit card fraud data},
  author={Leevy, Joffrey L and Hancock, John and Khoshgoftaar, Taghi M},
  journal={Journal of Big Data},
  volume={10},
  number={1},
  pages={118},
  year={2023},
  publisher={Springer}
}

@article{lemaavztre2017imbalanced,
  title={Imbalanced-learn: A python toolbox to tackle the curse of imbalanced datasets in machine learning},
  author={Lema{\~A}{\v{Z}}tre, Guillaume and Nogueira, Fernando and Aridas, Christos K},
  journal={Journal of machine learning research},
  volume={18},
  number={17},
  pages={1--5},
  year={2017}
}

@article{morteza2022novel,
  title={A Novel Gaussian-Copula modeling for image despeckling in the shearlet domain},
  author={Morteza, Arian and Amirmazlaghani, Maryam},
  journal={Signal Processing},
  volume={192},
  pages={108340},
  year={2022},
  publisher={Elsevier}
}

@article{paszke2019pytorch,
  title={Pytorch: An imperative style, high-performance deep learning library},
  author={Paszke, Adam and Gross, Sam and Massa, Francisco and Lerer, Adam and Bradbury, James and Chanan, Gregory and Killeen, Trevor and Lin, Zeming and Gimelshein, Natalia and Antiga, Luca and others},
  journal={Advances in neural information processing systems},
  volume={32},
  year={2019}
}

@inproceedings{patki2016synthetic,
  title={The synthetic data vault},
  author={Patki, Neha and Wedge, Roy and Veeramachaneni, Kalyan},
  booktitle={2016 IEEE international conference on data science and advanced analytics (DSAA)},
  pages={399--410},
  year={2016},
  organization={IEEE}
}

@article{platzer2021holdout,
  title={Holdout-based empirical assessment of mixed-type synthetic data},
  author={Platzer, Michael and Reutterer, Thomas},
  journal={Frontiers in big Data},
  volume={4},
  pages={679939},
  year={2021},
  publisher={Frontiers Media SA}
}

@inproceedings{roy2024frauddiffuse,
  title={FraudDiffuse: Diffusion-aided Synthetic Fraud Augmentation for Improved Fraud Detection},
  author={Roy, Ruma and Tiwari, Darshika and Pandey, Anubha},
  booktitle={Proceedings of the 5th ACM International Conference on AI in Finance},
  pages={90--98},
  year={2024}
}

@inproceedings{sattarov2023findiff,
  title={Findiff: Diffusion models for financial tabular data generation},
  author={Sattarov, Timur and Schreyer, Marco and Borth, Damian},
  booktitle={Proceedings of the Fourth ACM International Conference on AI in Finance},
  pages={64--72},
  year={2023}
}

@article{song2020score,
  title={Score-based generative modeling through stochastic differential equations},
  author={Song, Yang and Sohl-Dickstein, Jascha and Kingma, Diederik P and Kumar, Abhishek and Ermon, Stefano and Poole, Ben},
  journal={arXiv preprint arXiv:2011.13456},
  year={2020}
}

@article{sundaravadivel2025optimizing,
  title={Optimizing credit card fraud detection with random forests and SMOTE},
  author={Sundaravadivel, P and Isaac, R Augustian and Elangovan, D and KrishnaRaj, D and Rahul, VV Lokesh and Raja, R},
  journal={Scientific Reports},
  volume={15},
  number={1},
  pages={17851},
  year={2025},
  publisher={Nature Publishing Group UK London}
}

@article{tedesco2023gaussian,
  title={Gaussian copula modeling of extreme cold and weak-wind events over Europe conditioned on winter weather regimes},
  author={Tedesco, Paulina and Lenkoski, Alex and Bloomfield, Hannah C and Sillmann, Jana},
  journal={Environmental Research Letters},
  volume={18},
  number={3},
  pages={034008},
  year={2023},
  publisher={IOP Publishing}
}

@inproceedings{thanh2020catastrophic,
  title={Catastrophic forgetting and mode collapse in GANs},
  author={Thanh-Tung, Hoang and Tran, Truyen},
  booktitle={2020 international joint conference on neural networks (ijcnn)},
  pages={1--10},
  year={2020},
  organization={IEEE}
}

@article{vaswani2017attention,
  title={Attention is all you need},
  author={Vaswani, Ashish and Shazeer, Noam and Parmar, Niki and Uszkoreit, Jakob and Jones, Llion and Gomez, Aidan N and Kaiser, {\L}ukasz and Polosukhin, Illia},
  journal={Advances in neural information processing systems},
  volume={30},
  year={2017}
}

@article{villaizan2025diffusion,
  title={Diffusion models for tabular data imputation and synthetic data generation},
  author={Villaiz{\'a}n-Vallelado, Mario and Salvatori, Matteo and Segura, Carlos and Arapakis, Ioannis},
  journal={ACM Transactions on Knowledge Discovery from Data},
  volume={19},
  number={6},
  pages={1--32},
  year={2025},
  publisher={ACM New York, NY}
}

@article{xu2019modeling,
  title={Modeling tabular data using conditional gan},
  author={Xu, Lei and Skoularidou, Maria and Cuesta-Infante, Alfredo and Veeramachaneni, Kalyan},
  journal={Advances in neural information processing systems},
  volume={32},
  year={2019}
}

@article{zhang2025temporal,
  title={Temporal latent diffusion model for machine degradation trend forecasting},
  author={Zhang, Tian and Li, Hao and Jiao, Jinyang and Lin, Jing},
  journal={Knowledge-Based Systems},
  pages={114753},
  year={2025},
  publisher={Elsevier}
}

@inproceedings{zhao2021ctab,
  title={Ctab-gan: Effective table data synthesizing},
  author={Zhao, Zilong and Kunar, Aditya and Birke, Robert and Chen, Lydia Y},
  booktitle={Asian conference on machine learning},
  pages={97--112},
  year={2021},
  organization={PMLR}
}

@article{zhi2025research,
  title={Research on Modeling of the Imbalanced Fraudulent Transaction Detection Problem Based on Embedding-Aware Conditional GAN},
  author={Zhi, Luping and Wang, Wanmin},
  journal={Big Data Research},
  pages={100557},
  year={2025},
  publisher={Elsevier}
}

@article{zhu2024enhancing,
  title={Enhancing credit card fraud detection a neural network and smote integrated approach},
  author={Zhu, Mengran and Zhang, Ye and Gong, Yulu and Xu, Changxin and Xiang, Yafei},
  journal={arXiv preprint arXiv:2405.00026},
  year={2024}
}

\end{document}